\documentclass[final,5p,times,twocolumn]{elsarticle}

 \usepackage{graphicx}
 \graphicspath{ {./images/} }

\usepackage{amssymb}
\usepackage[dvipsnames]{xcolor}
\usepackage{enumerate}
\journal{Neurocomputing}

\begin{document}

\begin{frontmatter}



\title{Hybrid Channel Based Pedestrian Detection}


\author[mymainaddress]{Fiseha B. Tesema}\ead{falmikena@yahoo.com}
\author[address2]{Hong Wu
\corref{mycorrespondingauthor}}
\cortext[mycorrespondingauthor]{Corresponding author}\ead{hwu@uestc.edu.cn} 
\author[address2]{Mingjian Chen}\ead{starvingnow@163.com}
\author[address2]{Junpeng Lin}\ead{thanks3q@gmail.com}
\author[mymainaddress]{William Zhu}\ead{wfzhu@uestc.edu.cn}
\author[XJTLUaddress,AZFT]{Kaizhu Huang}\ead{kaizhu.huang@xjtlu.edu.cn}
\address[mymainaddress]{Institue of Fundamental and Frontier Science, University of Electronic Science and Technology of China, Chengdu, China}
\address[address2]{School of Computer Science and Engineering, University of Electronic Science and Technology of China, Chengdu, China}
\address[XJTLUaddress]{Department of Electrical and Electronic Engineering, Xi'an Jiaotong-Liverpool University, Suzhou, China}
\address[AZFT]{Alibaba-Zhejiang University Joint Institute of Frontier Technologies, Hangzhou, China}

\begin{abstract}
Pedestrian detection has achieved great improvements with the help of Convolutional Neural Networks (CNNs). CNN can learn high-level features from input images, but the insufficient spatial resolution of CNN feature channels (feature maps) may cause a loss of information, which is harmful especially to small instances. In this paper, we propose a new pedestrian detection framework, which extends the successful RPN+BF framework to combine handcrafted features and CNN features. RoI-pooling is used to extract features from both handcrafted channels (e.g. HOG+LUV, CheckerBoards or RotatedFilters) and CNN channels. Since handcrafted channels always have higher spatial resolution than CNN channels, we apply RoI-pooling with larger output resolution to handcrafted channels to keep more detailed information. Our ablation experiments show that the developed handcrafted features can reach better detection accuracy than the CNN features extracted from the VGG-16 net, and a performance gain can be achieved by combining them. Experimental results on Caltech pedestrian dataset with the original annotations and the improved annotations demonstrate the effectiveness of the proposed approach. When using a more advanced RPN in our framework, our approach can be further improved and get competitive results on both benchmarks.

\end{abstract}

\begin{keyword}
Pedestrian detection\sep Handcrafted features channels\sep CNN feature channels\sep RoI-pooling\sep Feature combination


\end{keyword}

\end{frontmatter}



\section{Introduction}
Pedestrian detection is a prerequisite task for many vision applications such as video surveillance \citep{surveillance2014, ian2019}, car safety \citep{autodriving2012}, and robotics. During the last decade, pedestrian detection has been attracting intensive research interests and great progress has been achieved \citep{benenson2014ten,toreaching}. Feature extraction is an important step for pedestrian detection algorithms, and various features have been proposed to improve detection accuracy. These image features can be grouped into handcrafted features and Convolutional Neural Network (CNN) features.

Traditional pedestrian detection methods employ the sliding window paradigm based on handcrafted features and traditional classifiers \citep{modelingdata2008}. Among the handcrafted features, the histogram of oriented gradient (HOG) descriptor \cite{dalal2005histograms} is the most well-known and works very well with linear Support Vector Machine (SVM) \citep{dalal2005histograms}. Felzenszwalb et al. \citep{5255236} proposed the Deformable Part Model (DPM) based on HOG features to handle pose variations of objects. Dollar et al. \citep{dollar2009integral} proposed Integral Channel Feature (ICF) which uses integral images to extract features from HOG channels and LUV color channels (HOG+LUV) and adopts boosted decision forests for pedestrian detection. Following ICF, further improvements have been made by applying various handcrafted or learned filters to the HOG+LUV channels, such as Aggregated Channel Features (ACF) \citep{6714453}, SquaresChntrs \citep{Benenson2013SeekingTS}, InformedHaar \citep{zhang2014informed}, Locally Decorrelated Channel Features (LDCF) \citep{nam2014local}, Checkerboards \citep{zhang2015filtered}, and RotatedFilters \citep{DBLP:journals/corr/ZhangBOHS16}. Extended Filtered Channel Framework (ExtFCF) \citep{8310009} combines several advanced filtered channel features (LDCF, Checkerboards, and RotatedFilters) in a  multilayer architecture to further improve performance. While these filtered channel feature detectors have achieved competitive results with low computational complexity, the handcrafted features are not robust enough for pedestrian detection in complex scenes.

Recently, deep CNNs have achieved great success in general object detection \citep{girshick2014rich,DBLP:journals/corr/HeZR014,girshick2015fast,ren2015faster,DBLP:journals/corr/RedmonDGF15,DBLP:journals/corr/LiuAESR15}. CNNs can learn high-level semantic features from input images, which are more discriminative for classification and have also been applied to pedestrian detection.  Early CNN-based pedestrian detection methods \citep{DBLP:journals/corr/HosangOBS15,DBLP:journals/corr/TianLWT14,tian2015deep,li2017scale} use handcrafted-feature-based detectors to generate a sparse set of pedestrian proposals and refine them with CNN classifiers. Some works \citep{yang2015convolutional}\citep{zhang2016faster}\citep{hu2018pushing} combine the feature maps from a pre-trained CNN model and a boosted forest classifier to improve the accuracy. The work \citep{zhang2016faster} also reveals that the coarse resolution of feature maps makes it hard to handle small instances. Later, more and more research works focus on developing end-to-end approaches by using customized architectures or adapting Fast-RCNN or Faster-RCNN to pedestrian detection. Scale Aware Fast R-CNN (SA-FastRCNN) \citep{li2017scale} extends Fast-RCNN with multiple built-in sub-networks to adaptively detect pedestrians of different scales. Multi-Scale CNN (MS-CNN) \citep{DBLP:journals/corr/CaiFFV16} and Multi-branch and High-level semantic convolutional neural Network (MHN) \citep{DBLP:journals/corr/abs-1804-00872} use feature maps of multiple layers to handle objects of different scales. TLL-TFA \citep{DBLP:journals/corr/abs-1807-01438} integrates somatic Topological Line Localization (TLL) network and Temporal Feature Aggregation (TFA) to detect multi-scale pedestrians. Graininess-aware Deep Feature Learning method (GDFL) \citep{Lin_2018_ECCV} uses scale-aware pedestrian attention masks and a zoom-in-zoom-out module to identify small and occluded pedestrians. Fused Deep Neural Network (F-DNN) \citep{du2017fused} combines multiple deep classifiers with a soft-reject strategy to refine proposals. Some other works improve pedestrian detection by using semantic segmentation as a strong cue \citep{du2017fused, DBLP:journals/corr/MaoXJC17, brazil2017illuminating, DBLP:journals/corr/abs-1805-08688} or part-based detection \citep{wang2018pcn}. While CNN-based approaches have consistently improved the detection accuracy, they always result in more complex architectures and higher computational costs.

Instead of designing new CNNs for pedestrian detection, a straightforward way to improve pedestrian detection is to combine the high resolution and easy-to-compute handcrafted feature channels with the low resolution and computational expensive CNN feature channels. Other than using handcrafted features to generate proposals and CNNs to classify them, some other methods have been proposed to take advantage of both kinds of features. Complexity Aware Cascade Training algorithm (CompACT) \citep{DBLP:journals/corr/CaiSV15} learnes a complexity aware cascade to integrate handcrafted features and CNN features with a trade-off between detection accuracy and speed. In Multi-layer Channel Features (MCF) \citep{7912366}, a multi-stage cascade AdaBoost is trained based on HOG+LUV and each layer of a CNN one after another. Mao et al. \citep{DBLP:journals/corr/MaoXJC17} integrated extra feature channels into CNN-based detectors by creating a new shallow branch alongside the back-bone CNN, but their experiments indicate that while the semantic features generated by a separated neural network can bring improvement to pedestrian detection, handcrafted features make almost no contribution. They further proposed Hyper-learner to jointly learn pedestrian detection as well as extra features supervised by semantic channels.

We follow this line and propose a simple and effective pedestrian detection framework called Hybrid Channel Detector (HCD) by extending the successful RPN+BF framework \citep{zhang2016faster} to combine handcrafted features and CNN features. For each proposal generated by RPN, ROI-pooling is used to extract features from both handcrafted feature channels and CNN feature channels, and with larger output resolution for handcrafted feature channels to retain more detailed information. Finally, the two kinds of features are concatenated and sent to boosted forests classifier. Overall, this work makes the following major contributions:
\begin{enumerate}[1)]
\item A simple yet effective framework is proposed to combine handcrafted features and CNN features for pedestrian detection. RoI-pooling is used to extract features from both handcrafted channels and CNN channels, and larger pooling output resolution is used for handcrafted channels to keep more detailed information.
\item We explore several handcrafted channels such as HOG+LUV, Checkerboards, and RotatedFilters in our framework. Our ablation experiments show that the handcrafted features can achieve better detection accuracy than the CNN features extracted from the VGG-16 net.
\item Experiments are conducted on Caltech pedestrian dataset \citep{5975165} with the original annotations and also a new set of improved annotations provided by \citep{DBLP:journals/corr/ZhangBOHS16}. Our approach achieves the MR of 9.53\% and 6.41\% on the two benchmarks respectively, both of which outperform previous approaches that combining handcrafted features and CNN features by a large margin. 
\item We further improve our approach by using SDS-RPN \citep{brazil2017illuminating}, a more advanced RPN, in our framework. And the approach achieves the MR of 8.62\% and 6.14\% on the two benchmarks respectively.
\end{enumerate}

The rest of the paper is organized as follows. Under Section II, we introduce some works related to our approach. Then, our proposed approach and its implementation details are introduced in Section III. Section IV shows the experimental results and analysis. Finally, we conclude this paper in Section V.

\begin{figure*}[t]
\includegraphics{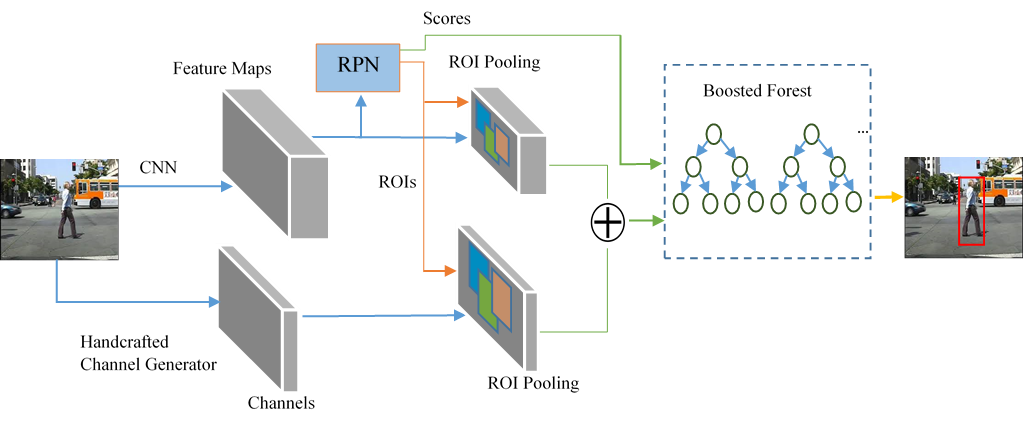}
\centering
\caption{
  The architecture of HCD: Top: RPN computes candidate bounding boxes and scores, and the backbone network generates CNN feature maps. Bottom: handcrafted feature channels are generated from the input image. Features are extracted by RoI-pooling from handcrafted feature channels and CNN feature maps respectively, and concatenated before being sent to Boosted forest (BF).}
  \label{arch1}
\end{figure*}

\section{Related works}
\subsection{ICF family pedestrian detectors}
In Integral Channel Features (ICF) \cite{dollar2009integral}, HOG+LUV channels are computed from an input image, and features such as local sums, histograms, and Haar features are extracted efficiently by using integral images. These features combined with boosted decision forests are very effective for pedestrian detection. To improve ICF, Aggregated Channel Feature (ACF) \citep{6714453} is proposed and becomes popular, in which features are single-pixel lookups from the aggregated (sum-pooled) channels. After that, various filtered channel features have been designed to compute over the HOG+LUV channels. The SquaresChntrs \citep{Benenson2013SeekingTS} includes a set of simple square pooling regions with different sizes. InformedHarr \citep{zhang2014informed} incorporates the statistical pedestrian model into the design of simple haar-like features. Nam et al. \citep{nam2014local} proposed LDCF whose filters are the top learned principal components analysis (PCA) eigenvectors. Zhang et al. \citep{zhang2015filtered} proposed a framework of filtered channel features to unify the aforementioned channel features, and systematically explored different filter banks. They found that the simple Checkerboards filters achieved the best performance among the filtered channel features. After that, Zhang et al. \citep{DBLP:journals/corr/ZhangBOHS16} proposed RotatedFilters, a simplification of Checkerboards inspired by LDCF and SquaresChntrs, generating a performance close to Checkerboards. Among these filtered channel features, LDCF is the only method whose filters are obtained by unsupervised learning rather than manually designed.

\subsection{Region-CNN for pedestrian detection}
The CNN-based object detectors can be classified to two groups, one-stage detectors \citep{DBLP:journals/corr/RedmonDGF15,DBLP:journals/corr/LiuAESR15} and two-stage detectors\citep{girshick2015fast, DBLP:journals/corr/HeZR014, ren2015faster, zhang2015filtered}. Although the one-stage detectors are more efficient than the two-stage ones, the current state-of-the-art object detectors are two-stage ones. In the two-stage frameworks, the first stage generates a sparse set of category-agnostic object proposals, and the second stage classifies the proposals into one of the foreground classes or background as well as refine their coordinates. 

Among the two-stage detectors, Region-CNN (R-CNN) family \citep{girshick2015fast, DBLP:journals/corr/HeZR014, ren2015faster, zhang2015filtered} is well-known and has consistently promoted the accuracy of object detection on major challenges and benchmarks. In R-CNN~\citep{girshick2014rich}, the selective search method is used to generate category-agnostic proposals, CNN is used to extract a fixed-length feature vector for each proposal, and an SVM classifier is used to classify the proposals. The speed of R-CNN is limited by the CNN feature extraction for each proposal. In Spatial Pyramid Pooling Network (SPPNet) \citep{DBLP:journals/corr/HeZR014}, spatial pyramid pooling is introduced to allow the computation of CNN feature extraction once per image. Based on SPPNet, Fast R-CNN \citep{girshick2015fast} uses the RoI-pooling and multi-task learning of class classification and bounding box regression and is trained end-to-end. Furthermore, Faster R-CNN \citep{ren2015faster} introduces a Region Proposal Network (RPN) that shares full-image CNN features with the detection network to efficiently generate object proposals. Since there is only one foreground object (pedestrian) in pedestrian detection, RPN can be trained as a pedestrian detector. In the work \citep{zhang2016faster}, it is found that RPN as a stand-alone pedestrian detector outperformes all handcrafted-feature-based detectors, and combining RPN with a boosted forest classifier further improves the performance.

\section{Hybrid channel detector}

The handcrafted channel features, like ICF and other filtered channel features, are able to describe the low-level and even middle-level image information well and have been proven effective for pedestrian detection. Meanwhile, CNNs can learn not only low-level and middle-level features in its first few layers, but also high-level features in the last few layers. While CNN feature channels (feature maps) get lower resolutions in higher layers, the HOG+LUV channels and other filtered channels have the resolutions close to the original image and can keep more detailed information. To take advantage of both sides, we combine handcrafted channels and CNN channels to construct a hybrid channel detector (HCD).

The architecture of our hybrid channel detector is given in Figure \ref{arch1}. It consists of four basic modules: 1) Firstly, pedestrian proposals are produced by a proposal generation network, e.g. RPN, and CNN feature channels are generated by the back-bone CNN. 2) Secondly, traditional handcrafted channels, like HOG+LUV or other filtered channels, are generated from an input image, and this can be done in parallel with the first step. 3) Thirdly, for each pedestrian proposal, RoI-pooling is used to extract the feature vectors from the two kinds of feature channels respectively. 4) Finally, the two kinds of features are concatenated and sent to a boosted forest classifier. 

\begin{figure*}[t]
\includegraphics{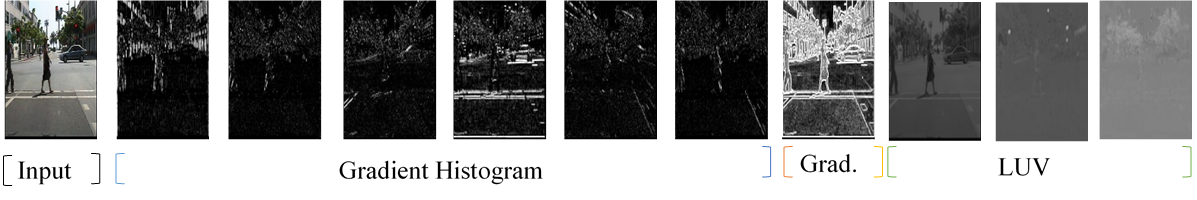}
\centering
\caption{
  Example image and computed HOG+LUV channels}
  \label{hogluv}
\end{figure*}

\subsection{Region proposal network}
Faster R-CNN \citep{ren2015faster} is a successful method for general object detection. It consists of two components: a Region Proposal Network (RPN) for generating region proposals, followed by a Fast R-CNN \citep{girshick2015fast} to classify the proposal regions into object categories or background. The RPN shares full-image CNN features with the Fast R-CNN, thus enabling nearly cost-free region proposals. The RPN is composed of an intermediate $3\times3$ convolutional layer and followed by two sibling $1\times1$ convolutional layers for classification and bounding box regression. The RPN uses classifiers and regressors on sliding windows over CNN feature maps to predict proposals relative to anchor boxes of multiple scales and aspect ratios (3 scales and 3 aspect ratios used in \citep{ren2015faster}). The classifier determines the probability of a proposal having an object, and the regressor regresses the coordinates of the proposal. The cost-function for training RPN contains both classification loss and regression loss.

RPN is a class-agnostic proposal generator in the scenario of multi-category object detection. For single-class object detection, RPN is naturally a detector for the only foreground category concerned. In \citep{zhang2016faster}, RPN is specially tailored for pedestrian detection and achieves competitive results as a stand-alone pedestrian detector. VGG-16 net \citep{DBLP:journals/corr/SimonyanZ14a} pre-trained on ImageNet dataset \citep{DBLP:journals/corr/RussakovskyDSKSMHKKBBF14} is adopted as the backbone network. Instead of 3 aspect ratios used in the original RPN, a single aspect ratio of 0.41 (width to height) is used, since it is the average aspect ratio of pedestrians as indicated in \citep{5975165}, and anchors of inappropriate aspect ratios would be noisy and harmful to detection accuracy. In addition, anchors of 9 different scales are used, starting from 40 pixels height with a scaling stride of 1.3$\times$, which spans a wider range of scales than the original RPN. With these adaptations, RPN achieves the log-average miss-rate (MR) of 14.9\% on the Caltech evaluation benchmark, which outperformes the leading traditional pedestrian detection methods: SCF \citep{benenson2014ten}, LDCF \citep{nam2014local}  and Checkerboards \citep{zhang2015filtered}. In this paper, we follow the adaptations for RPN in \citep{zhang2016faster}. 

In fact, any proposal generation network can be used in place of RPN in our framework. Simultaneous Detection Segmentation RCNN (SDS-RCNN) \cite{brazil2017illuminating} contains a region proposal network to generate pedestrian proposals and a Binary Classification Network (BCN) to classify them. Box-based semantic segmentation is introduced to both networks as auxiliary tasks. SDS-RCNN achieves leading performance on Caltech dataset, and even its region proposal network, SDS-RPN, achieves promising 9.63\% miss rate. In our experiments, we also evaluate our framework with SDS-RPN to demonstrate its generality.

\subsection{ Handcrafted feature channels}

In traditional pedestrian detection, an image is first transformed into the 10 HOG+LUV channels (shown in figure \ref{hogluv}), which include normalized gradient magnitude, histogram of oriented gradients on 6 directions, and LUV color channels. Then different filtered channel features \citep{zhang2015filtered} have been proposed by convolving different filter banks with the HOG+LUV channels. Among these filtered channel features, the simple Checkerboards filters  \citep{zhang2015filtered} achieves the best performance. Zhang et al. \citep{DBLP:journals/corr/ZhangBOHS16} further proposed RotatedFilters, a simplification of Checkerboards. Therefore, we investigate to apply the basic HOG+LUV, Checkerboards, and RotatedFilters to our framework in this paper.

\begin{figure}
  \centering
\includegraphics[width=5cm,height=5cm,keepaspectratio,]{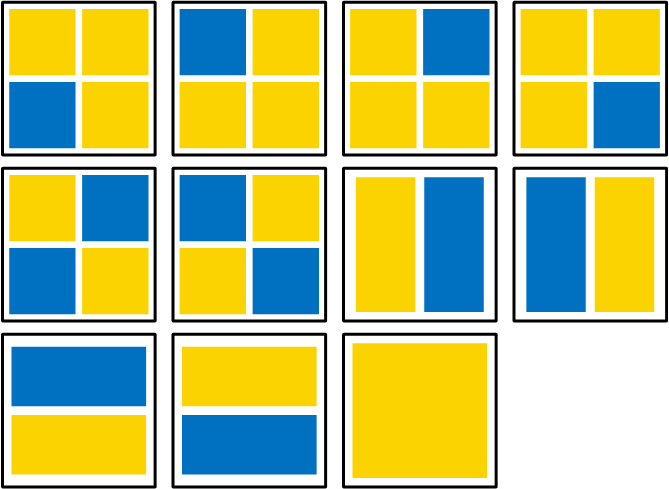}

\caption{
Eleven 2 $\times$ 2 Checkerboards-like filters. Blue or Yellow represents
+1 or -1, respectively.
  }
  \label{ch11}
\end{figure}

\textbf{Checkerboards:} Checkerboards is a na\"ive set of filters of the same sizes: a uniform square, all horizontal and vertical gradient detectors ( $\pm1$ values), and all possible checkerboard patterns. The number of filters increases rapidly with checkerboard size. For up to $4\times4$ cells, there are 61 filters (CB61). But, more filters lead to higher computaional complexity for training and testing. Cai et al.\cite{DBLP:journals/corr/CaiSV15} adopted eight $2\times2$ Checkerboards-like filters for limited complexity. You et al. \citep{8310009} added three more homologous filters to get 11 filters (Checkerboards-11, CB11) in total, as shown in Figure \ref{ch11}. After applying the 11 filters to the 10 HOG+LUV channels, 110 feature channels are finally generated. CB11 achieves competitive performance (MR = 18.60\%) on Caltech compared to the original CB61 (MR = 18.47\%), and run $\sim5\times$ faster than it. In this paper, we adopt CB11 for its good trade-off between the accuracy and complexity.

\begin{figure}
  \centering
\includegraphics[width=\linewidth]{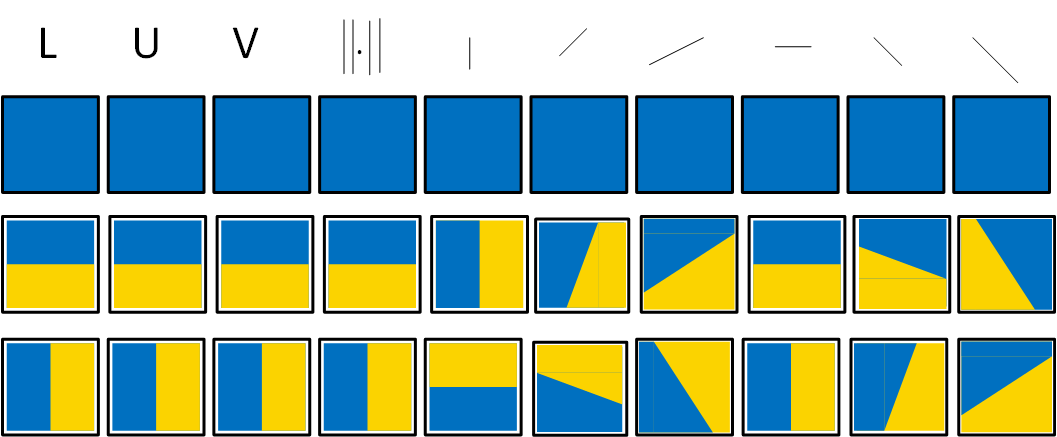}
\caption{
Filter patterns of RotatedFilters. Blue or Yellow represents +1 or -1 respectively.
  }
  \label{RF9}
\end{figure}

\textbf{RotatedFilters:} RotatedFilters (RF) \citep{DBLP:journals/corr/ZhangBOHS16} (shown in Figure \ref{RF9}) is inspired by the filterbank of LDCF \citep{nam2014local}. The first three filters for each feature channel of HOG+LUV are the constant filter and two step functions in orthogonal directions, with the two step functions rotated for the oriented gradient channels of HOG. In the same spirit as SquaresChnFtrs \citep{Benenson2013SeekingTS}, each filter per channel repeats over 3 scales ($4\times4$, $8\times8$, and $16\times16$), resulting in 9 filters per channel. After applying these filters to the 10 HOG+LUV channels, 90 feature channels are finally generated. RotatedFilters also achieves competitive performance (MR = 19.20\%) compared to CB61, and run $\sim6\times$ faster than it. RF is also explored in our study. The main difference between RF and CB11 is that CB11 has abundant filters whereas RotatedFilters focuses on fine and coarse grained information under the multi-scale design.

\subsection{ Feature extraction and fusion}
By now, both the handcrafted channels and the CNN channels have been computed from an input image, and the proposals generated by RPN need to be classified based on the feature vectors extracted from the two kinds of feature channels.

RoI-pooling is a feature extraction method proposed in \citep{girshick2015fast}, which extracts a fixed-length feature vector from feature maps for each object proposal. It can achieve a significant speedup of both training and testing and also maintain a high detection accuracy. Thus, it has been widely used in CNN-based object detection. The RoI-pooling layer uses max-pooling to convert the features inside any valid region of interest (RoI) into a small feature map with a fixed spatial extent of $H\times W$ (e.g., $7\times7$). RoI-pooling works by dividing an $h\times w$ input RoI window into a $H\times W$ grid of sub-windows of approximate size $h/H \times w/W$ and then max-pooling the values in each sub-window into the corresponding output grid cell. If an RoI-pooling's input resolution is smaller than the output resolution (e.g., $<7\times 7$), the pooling bins collapse and the features become less discriminative. In \citep{zhang2016faster}, the same RoI-pooling output resolution ($7\times 7$) as in Faster R-CNN \citep{ren2015faster} is used to extract features from feature maps of different layers, and Conv3\_3 gives the best accuracy among them. Since the feature maps in the 3rd layer has a 4$\times$ downsampling factor compared to the input image, the features of object proposals less than $28\times28$ in an input image are less discriminative. If the features are extracted from Conv5\_3 which has a 16$\times$ downsampling factor, the features of the object proposals less than $112\times112$ in an input image are less discriminative, and not suitable for small object detection. This would be an important reason that the features from Conv3\_3 is remarkably better than those from Conv5\_3. In \citep{zhang2016faster}, the problem of low-resolution feature maps is alleviated by extracting features from lower layer (Conv3\_3), and by using the \`{a} trous trick [16] to increase feature map resolution. In this paper, RoI-pooling with output resolution $7\times7$ is used to extract features from Conv3\_3.

For handcrafted channels, we also use RoI-pooling to extract a feature vector for each proposal. Handcrafted channels, like HOG+LUV and the filtered channels, have the same or similar resolution as the input image. Therefore we can use RoI-pooling with larger output resolution. When the output resolution gets larger, the extracted features can retain more detailed information but have a larger dimension. In this paper, we investigate several output resolutions ($7\times7$, $14\times14$, $20\times20$ and $28\times28$) for handcrafted feature extraction. After feature vectors extracted from the two kinds of feature channels, they are concatenated and sent to the classifier. This kind of feature combination allows the two kinds of features to have different lengths. 

\subsection{Boosted forest}

Boosted Forest (BF) or Boosted Decision Trees (BDT) is an ensemble learning method. It trains a number of decision trees sequentially with a boosting process. Boosting re-weights versions of the training data depending on whether the previously trained tree classifies them correctly, and makes the final decision by taking a weighted majority vote of the sequence of classifiers produced so far \citep{article}. Boosted Forest can achieve fast and accurate classification, and has been widely used in computer vision such as object recognition \cite{wohlhart2012discriminative,gall2013class}, pedestrian detection \citep{dollar2009integral,zhang2016faster,6751366,6714453,nam2014local,yang2015convolutional} and super-resolution \cite{huang2015fast, huang2017learning,huang2017srhrf+}. In \citep{zhang2016faster}, BF classifier is applied to RPN proposals for pedestrian detection, and is superior to Fast R-CNN classifier in the following aspects. Firstly, BF is extremely light-weight and easy to learn with fewer hyper-parameters; secondly, it is less prone to over-fitting and adversarial examples \citep{szegedy2013intriguing, yang2015convolutional, KHuangICDM2015,MAT-KHuang2018}; thirdly, the BF is flexible for combining features with various dimensions; finally, it is flexible for incorporating effective bootstrapping for mining hard negatives. Owing to these merits of Boosted Forest, we adopt it to classify proposals in our framework. 

\subsection{Implementation details}
The program for computing all the handcrafted channels, including HOG+LUV, CB11, and RF, is implemented based on Piotr Dollar's Matlab toolbox \citep{pdollar}. The source codes of \citep{zhang2016faster} are used for implementing RPN and BF in our framework and the training procedure and related hyper-parameter settings in \citep{zhang2016faster} are also followed. VGG-16 net \citep{DBLP:journals/corr/SimonyanZ14a} pre-trained on the ImageNet dataset \citep{DBLP:journals/corr/RussakovskyDSKSMHKKBBF14} is used as the backbone network, and RPN is fine-tuned to Caltech dataset in the same way as in \citep{zhang2016faster}. After proposal generation, non-maximum suppression (NMS) with a threshold of 0.7 is also used to filter the proposal regions. Then the proposal regions are ranked by their scores, where the top-ranked $1,000$ proposals (and ground truths) of each image are used for training BF classifier, and the top-ranked 100 proposals are used for testing. The RealBoost algorithm is used for BF classifier. The training is bootstrapped for 6 times, and after each stage, hard negative mining is also applied. The forest in each stage consists $\left\{64, 128, 256, 512, 1024, 1536\right\}$ trees, and tree depth is set to 5. After the 6 bootstrapping stages, a forest of $2,048$ trees is trained and used for inference. Following \citep{zhang2016faster}, the confidence score of each proposal computed by RPN is used to compute the stage-0 classifier. We also adopt single-scale training and testing, without using feature pyramids. An image is resized such that its shorter edge has 720 pixels before training and testing. For more detailed information, please refer to \citep{zhang2016faster} and its publicly available source codes. For SDS-RPN\cite{brazil2017illuminating}, we simply use the trained model provided by the authors.

\section{Experiments and analysis}
\subsection{Datasets and evaluation metrics}
We evaluate the effectiveness of our approach on Caltech dataset \citep{5975165}, and also evaluate our framework with SDS-RPN to demonstrate its generality. The evaluation metric employed is the log average Miss Rate on False Positive Per Image (FPPI) in [10$^{-2}$, 10$^{0}$], denoted as MR\_2, or in short MR. Caltech pedestrian dataset and its associated benchmark are widely-used for evaluation of pedestrian detection. The dataset is comparatively large and challenging, consisting of about 10 hours of videos (30 frames per second) collected from a vehicle driving through urban traffic. Every frame in the raw Caltech dataset has been densely annotated with the bounding boxes of pedestrian instances. There are totally $350,000$ bounding boxes of about $2,300$ unique pedestrians labeled in $250,000$ frames. Various evaluation settings have been defined based on height and fraction of occlusion of pedestrians. The settings used in this paper are introduced as follows.

\textbf{Reasonable subset:} Pedestrian instances have 50 or more pixels in height and at least 65\% visible body parts. It is the most frequently used evaluation setting and considered as a standard evaluation benchmark in almost all pedestrian detection research works. In this paper, we also use it as the default setting for evaluation.

\textbf{Occlusion-related settings:} Partial and heavy occlusion subsets include pedestrians taller than 50 pixels and having 1\%--35\% and 36\%--80\% body parts occluded respectively. 

\textbf{Scale-related settings:}
Near and Medium subsets include pedestrians have 80 or more pixels and 30-80 pixels in height respectively.

In addition to Caltech dataset with the original annotations, we also evaluate our approaches on the new annotations provided by \citep{DBLP:journals/corr/ZhangBOHS16}, which corrects the errors in the original annotations. For simplicity, RPN and SDS-RPN are not fine-tuned to the new annotations when evaluating on them.

\subsection{Evaluation of handcrafted channels}
We first carry out ablation experiments to evaluate the handcrafted channel features. That is, we only use the features extracted from the handcrafted channels, and do not use the features from the CNN channels. Since the handcrafted channels have higher resolution than the CNN channels, RoI-pooling with higher output resolutions can be used to retain more detailed information. In the experiments, we explore four kinds of RoI-pooling output resolutions, $7\times7$, $14\times14$, $20\times20$ and $28\times28$, and compare three handcrafted channels, HOG+LUV, CB11 and RF. The experimental results are shown in table \ref{compr_Fil}.

As shown in Table \ref{compr_Fil}, HOG+LUV, CB11, and RF achieve their best MR of 11.79\%, 11.12\% and 10.6\% respectively. Among them, RF achieves the best result, and the MR of HOG+LUV is only 1.2 percentage points lower than that. This is an amazing result, considering that HOG+LUV has only 10 channels while CB11 and RF have 110 and 90 channels respectively. All these performances are better than the best MR of 12.4\% achieved by CNN features from Conv3\_3 \citep{zhang2016faster}. Even with the same output resolution of $7\times7$ as in \citep{zhang2016faster}, CB11 and RF outperform Conv3\_3, and HOG+LUV outperforms Conv2\_3 and Conv5\_3. Since the RPN+BF framework is used in both our study and the work in \citep{zhang2016faster}, it can be concluded that the developed handcrafted channel features have better representation power than the CNN features of VGG-16 pre-trained on ImageNET. To the best of our knowledge, this observation has not been reported by previous works. Traditional methods always adopt the handcrafted features with the sliding-window paradigm, and the leading handcrafted channel feature CB61 only achieves MR of 18.47\%. This also indicates the RPN+BF framework is superior to the traditional sliding-window paradigm. HOG+LUV and RF achieve the best performance with the RoI-pooling output resolution of 20$\times$20, and CB11 achieves the best one with 28$\times$28. In the following experiments, we use the RoI-pooling output resolution of 20$\times$20 for all the three handcrafted channel features.

\begin{table}
  \caption{Detection performance (MR\%) of handcrafted channel features with different RoI-pooling resolutions. }
  \label{compr_Fil}
  \centering 
\begin{tabular}{l|l|l|l|l}\hline
     RoI features & 7$\times$7 & 14$\times$14 & 20$\times$20 & 28$\times$28 \\ \hline \hline
     HOG+LUV & 14.47 & 13.88 & \textbf{11.79}   & 12.81  \\
     CB11 & 12.34 & 11.21 & 11.51 & \textbf{11.12} \\
     RF & 12.09 & 10.94 & \textbf{10.6}  & 11.26    \\  \hline
\end{tabular}
\end{table}

\subsection{Evaluation of hybrid channels detector}

This section is devoted to exploring the effectiveness of the proposed hybrid channels detector. The feature maps from Conv3\_3 of VGG-16 are used as the CNN channels in our framework because it achieves the best performance among the different layers in \cite{zhang2016faster, yang2015convolutional}. HOG+LUV, CB11, and RF are used as the handcrafted channels respectively. For the CNN feature channels, the output resolution of RoI-pooling is set to $7\times7$, the same as in \cite{zhang2016faster, yang2015convolutional}. The features extracted from the two kinds of channels are concatenated before being sent to the BF classifier. 

Table \ref{Comp_HCD} shows the results of different combinations of handcrafted features and CNN features and also the results of the combinations of different CNN features in \cite{zhang2016faster}. From the results, we can see that, combination of handcrafted features and CNN features outperforms each of CNN features alone. They outperform the approaches based only on handcrafted channel features for about 1.5 percentage points and outperform the approaches based only on Conv3\_3 for about 2.5 percentage points. Among them, the combination of RF and Conv3\_3 achieves the best MR at 9.53\%, which is slightly better than the best MR of 9.6\% achieved in \cite{zhang2016faster}. In their work, the best result is achieved by the combination of Conv3\_3 and the \'{a} trous version of Conv4\_3, which takes extra computation time to re-compute the Conv4\_3 features maps with the \'{a} trous trick. While in our approach, the handcrafted channels can be computed in parallel with VGG-16. The combination of HOG+LUV and Conv3\_3 achieves the MR of 10.03\%, and the combination of CB11 and Conv3\_3 achieves the MR of 10\%. All our three kinds of combination outperform the combinations of CNN features from two or three layers of VGG-16 in \cite{zhang2016faster}. This indicates the handcrafted features can provide more complementary information to the CNN features. 

\begin{table}
  \caption{Detection performance of different combinations of channel features.}
  \label{Comp_HCD}
  \centering 
\begin{tabular}{l|l}\hline
    RoI features & MR(\%)\\ \hline \hline
     (HOG+LUV), Conv3\_3  &10.03 \\
     CB11, Conv3\_3  & 10.00  \\
     RF, Conv3\_3  & \textbf{9.53}\\ \hline
     Conv3\_3, Conv4\_3 \citep{zhang2016faster} &11.5\\  
     Conv3\_3, Conv4\_3, Conv5\_3 \citep{zhang2016faster} & 11.9 \\ 
     Conv3\_3, Conv4\_3 (\`{a} trous) \citep{zhang2016faster} & 9.6 \\\hline
\end{tabular}
\end{table}

 \subsection{Evaluation with respective to occlusion and scale} 
We further evaluate our approach under different scale evaluation settings (Near and Medium) and different occlusion evaluation settings (None, Partial and Heavy). Table \ref{table_scale} lists the results of RPN as a stand-alone detector, our framework with Conv3\_3, HOG+LUV, CB11 and RF alone respectively, and the combination of RF and Conv3\_3. All tested methods beat RPN stand-alone detector, and the combination of RF and Conv3\_3 achieves the best performance. All three handcrafted features outperform Conv3\_3 not only on the Near subset but also on the Medium subset. This indicates that the larger RoI output resolution can help handcrafted features to handle smaller objects. Under the occlusion-related settings, CB11 and Conv3\_3 slightly outperform RF on the Heavy and Partial occlusion subsets, but RF is far ahead on the None occlusion subset. RF gets very good results under the easier settings like Near and None-occlusion, and also outperforms Conv3\_3 under these settings. The combination of RF and Conv3\_3 can further improve the performance under all the settings, and more relative improvement can be made under the easier settings (e.g. Near and None occlussion). Comparing the combination of RF and Conv3\_3 with RPN stand-alone, while the former reducing 6.4\% of MR from 58.09\% to 54.35\% on the Medium subset, it reduces 35.9\% of MR from 2.59\% to 1.66\% on the Near subset. This may suggest that high-level features and/or more sophistic learning schemes are necessary to handle more difficult situations like small scale and heavy occlusions.
 
\begin{table}
  \caption{Detection performance (MR) under different scales and occlusion settings}
  \label{table_scale}
  \centering 
\begin{tabular}{l|l|l|l|l|l}\hline
     Methods  & Medi. & Near & Heavy  & Partial & None\\ \hline \hline
      RPN stand-alone & 58.09 & 2.59 & 78.77  & 27.64 &11.33\\
      HOG+LUV  & 56.31 & 2.59 & 76.62  & 26.54 &9.95\\
      CB11 & 56.11 & 2.54 & 75.08  & 24.80 &10.01\\ 
      RF  & 55.85 & 2.05 & 75.57  & 24.92 & 8.88\\
      Conv3\_3 & 56.61 & 3.9 & 75.08  & 24.91 & 10.07\\
      RF, Conv3\_3  & \textbf{54.35} & \textbf{1.66} & \textbf{74.72}  & \textbf{20.4} & \textbf{8.15} \\ \hline
\end{tabular}
\end{table}

\begin{table}
  \caption{Comparison of our works with other works on Caltech with the original and new annotations.}
  \label{Comp_Flb+Conv3}
  \centering 
\begin{tabular}{l|l|l}\hline
     Methods& MR$^O$ & MR$^N$\\ \hline \hline
%CCF+CF\citep{yang2015convolutional}  & 17.32 &-\\
CompACT-Deep \cite{DBLP:journals/corr/CaiSV15} & 11.75 & 9.15\\
MCF \citep{7912366} & 10.40 &7.98\\
SDS-RPN \cite{brazil2017illuminating} & 9.63 & -\\
RPN+BF \citep{zhang2016faster} & 9.6 & 7.3 \\
\textbf{HCD} & 9.53 & 6.41 \\
F-DNN \cite{DBLP:journals/corr/abs-1805-08688}& 8.65 & -\\
\textbf{HCD} (SDS-RPN) & 8.62 & 6.14\\
PCN \cite{wang2018pcn} & 8.4&-\\
F-DNN+SS \cite{DBLP:journals/corr/abs-1805-08688}& 8.18 & 6.89\\
GDFL \cite{Lin_2018_ECCV}& 7.85&-\\
TLL-TFA\cite{DBLP:journals/corr/abs-1807-01438} & 7.39 &-\\
SDS-RCNN \cite{brazil2017illuminating} & 7.36 & 6.44\\
HyperLearner\citep{DBLP:journals/corr/MaoXJC17}&-&5.5 \\
 \hline
\end{tabular}
\end{table}

\subsection {Comparison with state-of-the-art methods on Caltech with the original and new annotations}
Table \ref{Comp_Flb+Conv3} lists our best result (the combination of RF and Conv3\_3) and those of the state-of-the-art methods on Caltech dataset with original and new annotations. \textit{O} stands for the original annotations, and \textit{N} for the new ones. Our approach has achieved MR of 9.53\% and 6.41\% on the two benchmarks respectively. Both of the results are much better than the previous methods which combine handcrafted features and CNN features, e.g. CompACT \cite{DBLP:journals/corr/CaiSV15} and MCF \citep{7912366}. Our approach reaches the same results as previous RPN+BF \citep{zhang2016faster} on the original Caltech dataset, and gets a better result of 6.41\% over its 7.3\% on Caltech dataset with the improved annotations. 

When we use SDS-RPN \citep{brazil2017illuminating} in our framework for proposal generation, our approach can be improved further. SDS-RPN as stand-alone pedestrian detector achieves the MR of 9.63\% on Caltech dateset, and our approach achieves the MR of 8.62\% and 6.1\% on the two benchmarks respectively. These results demonstrate the generality of our framework.

There are some research works which have obtained better performance than ours, but they always have more complex architectures and higher computational costs. In the work \citep{DBLP:journals/corr/MaoXJC17}, the combination of CNN features and HOG+LUV channel features gives almost no improvement, and Hyper-learner is further proposed to jointly learn pedestrian detection as well as extra features supervised by semantic channels and achieves leading performance on Caltech dataset with the new annotations, but it needs another CNN to generate the semantic channels. Part and context network (PCN) \cite{wang2018pcn} consists of three DNNs subnetworks: a part branch which uses LSTM for semantic information communication, an original branch using an original box from RPN, and a context branch introducing max-out operation for region scale selection. F-DNN \citep{du2017fused} uses a soft-rejection to fuse multiple deep neural networks to classify the candidates. F-DNN+SS \cite{du2017fused} further uses a pixel-wise semantic segmentation network to refine the classification and improves accuracy at the expense of a significant loss in speed. GDFL \cite{Lin_2018_ECCV} includes three components: a convolutional backbone, a scale-aware pedestrian attention module and a zoom-in-zoom-out module to identify small and occluded pedestrians. TLL-TFA \cite{DBLP:journals/corr/abs-1807-01438} integrates the somatic topological line localization (TLL) networks and temporal feature aggregation for detecting multi-scale pedestrians. In SDS-RCNN \cite{brazil2017illuminating}, their RPN and Binary Classification Network (BCN) do not share features, and box-based semantic segmentation is introduced to both networks as auxiliary tasks. 

Since HOG can be accelerated by GPU implementation, and CheckerBoards and RotatedFilters can be implemented with convolution layers, our HCD can archive the efficiency close to the original RPN+BF. If the handcrafted channels are computed in parallel with RPN, HCD can reach the same speed as RPN+BF.

\section{Conclusion}
In this paper, we propose a novel hybrid channel based pedestrian detector which extends the RPN+BF framework to integrate handcrafted features and CNN features. Our ablation experiments reveal that the handcrafted features extracted by RoI-pooling with larger output resolution can achieve better detection accuracy than the CNN features from the VGG-16 net. With the combination of the efficient handcrafted features and high-level CNN features, our approach achieves comparable detection accuracies on the Caltech pedestrian dataset with the original and the improved annotations, and outperforms previous methods which combine handcrafted features and CNN features by a large margin. When we use SDS-RPN in our framework for proposal generation, our approach can be improved further. This also demonstrate the generality of our framework. Our study also indicates that handcrafted features are complementary to CNN features for pedestrian detection and it can be further explored in other deep frameworks.

\section*{Acknowledgment}
We greatly acknowledge support by the National Natural Science Foundation of China under Grant Nos. 61772120 and 61876155.


%%       \citet{<label>} ==> Jones et al. [21]
%%       \citep{<label>} ==> [21]
%%

\section*{References}
\bibliographystyle{elsarticle-num-names} 
\bibliography{handcrafted_channel_detector}


%\begin{thebibliography}{00}



%\end{thebibliography}
\end{document}